\documentclass[runningheads]{llncs}
\usepackage[T1]{fontenc}
\usepackage{graphicx}
\usepackage{booktabs}
\usepackage{amsmath}
\usepackage{amsfonts}
\usepackage{algorithm}
\usepackage{algorithmic}
\usepackage{float}
\usepackage{placeins}
\usepackage{microtype}
\usepackage{xurl}

\begin{document}

\title{BERTilda: Explainable Topic Lifecycle Tracking with Split/Merge Detection via Similarity-and-Flow Temporal Graphs}
\titlerunning{BERTilda}

\author{Cl\'audia Oliveira\inst{1}\orcidID{0009-0004-2789-0282} \and
\'Alvaro Figueira\inst{1,2}\orcidID{0000-0002-0507-7504}}
\authorrunning{C. Oliveira and \'A. Figueira}
\institute{Faculty of Sciences, University of Porto, Portugal \\
\email{arfiguei@fc.up.pt}
\and
INESCTEC, Porto, Portugal}

\maketitle

\begin{abstract}
Longitudinal text streams exhibit topic birth and death, but also discrete structural reorganizations in which themes \emph{split} into subtopics or \emph{merge} into broader narratives. Many dynamic topic models emphasize smooth drift, while snapshot topic models (fit independently per time window) leave temporal correspondence underspecified.
We present \emph{BERTilda}, an explainable framework that discovers topics independently in each window (using an embedding-based topic model) and then constructs a \emph{temporal topic graph} linking topics across adjacent windows. Links are supported by two complementary signals: (i) semantic similarity between topic representations and (ii) a \emph{bidirectional coverage} signal that estimates document outflow (where a topic goes) and inflow (where a topic comes from) via cross-window tweet-to-topic attribution.
Graph-based rules label continuations, splits, merges, disappearances, and unclear transitions.
We evaluate BERTilda on political corpora, including U.S. congressional tweets and historical speech datasets, report topic-quality and temporal-stability diagnostics, and validate lifecycle labels on a gold-standard subset annotated by three independent annotators. On the annotated subset, BERTilda reaches majority agreement rates up to 87\% and attains the highest macro-average agreement across the compared methods, with particularly strong disappearance detection relative to similarity-only and forward-only baselines.

\end{abstract}

\section{Introduction}
\label{sec:intro}

Topic modeling is a standard tool for exploring large text collections, yet many real-world corpora are inherently longitudinal. In political communication on social media and in institutional speeches, themes do not only drift gradually: topics are born and die, but also fragment into competing sub-narratives (splits) or converge when previously separate debates become tightly coupled (merges). Explicitly identifying such structural events is useful for summarizing how attention shifts and reorganizes over time.

Methodologically, split/merge detection is difficult because the concept of a \emph{topic} depends on the modeling family. Classical probabilistic topic models such as LDA \cite{10.5555/944919.944937} and many dynamic extensions (e.g., DTM \cite{10.1145/1143844.1143859}, Topics over Time \cite{10.1145/1150402.1150450}, and continuous-time variants \cite{wang2015continuoustimedynamictopic}) define topics as distributions over words and often assume relatively smooth evolution. Neural dynamic models (e.g., DETM \cite{dieng2019dynamicembeddedtopicmodel}) instead evolve latent representations. These formulations provide valuable temporal regularization, but they do not naturally yield discrete, auditable split/merge events. Conversely, snapshot topic models, including embedding-based approaches effective for short texts (e.g., Top2Vec \cite{DBLP:journals/corr/abs-2008-09470}, BERTopic \cite{grootendorst2022bertopicneuraltopicmodeling}, and embedding-space topic models \cite{dieng2020topicmodelinginembeddingspaces}), avoid imposing continuity but leave topic correspondence across windows ambiguous.

We propose BERTilda, "BERT-based Temporal Identification, Lifecycle Detection and Analysis", which treats temporal topic modeling as an alignment-and-event-labeling problem on an explicit graph. Topics are discovered independently within each window using a snapshot topic model (we instantiate with BERTopic \cite{grootendorst2022bertopicneuraltopicmodeling}, but the framework is model-agnostic). We then link topics across adjacent windows using two signals. The first is semantic similarity between topic representations in a shared embedding space. The second is \emph{bidirectional coverage}, an interpretable flow signal that estimates (i) document \emph{outflow} from a source topic and (ii) document \emph{inflow} into a target topic via cross-window document-to-topic attribution.
Together, we use these signals and apply transparent rules to label continuations, splits, merges, disappearances, and unclear transitions, which are then used to construct a temporal topic graph.

Empirically, we evaluate BERTilda on a corpus of 357,896 tweets authored by 544 members of the 119th U.S. Congress, and on historical political speech corpora used in prior dynamic topic modeling work. We report topic quality and temporal stability diagnostics and validate lifecycle labels on a gold-standard subset annotated by three independent annotators.

This paper makes three contributions: (i) bidirectional coverage as an interpretable flow signal for temporal topic alignment that complements embedding-based similarity; (ii) a graph-based labeling scheme that operationalizes topic lifecycles and split/merge events without assuming smooth evolution; and (iii) an evaluation protocol that combines quantitative diagnostics with human validation on political corpora.

The remainder of this paper is organized as follows. Section~\ref{sec:rw} reviews related work on topic models, temporal evolution, split/merge detection, and evaluation. Section~\ref{sec:method} presents BERTilda. Section~\ref{sec:exp} describes datasets, baselines, and evaluation protocols. Section~\ref{sec:results} reports results, followed by discussion and limitations in Section~\ref{sec:discussion}.

\section{Related Work}
\label{sec:rw}

\subsection{Topic modeling for short and noisy texts}
Early topic models represent documents as mixtures of latent components defined over vocabularies. Classical examples include probabilistic latent semantic analysis \cite{hofmann1999plsa} and LDA \cite{10.5555/944919.944937}. Nonnegative matrix factorization (NMF) provides an alternative matrix-decomposition perspective \cite{lee1999learningparts,lee2001algorithms}, while Bayesian nonparametric approaches such as hierarchical Dirichlet processes address model selection \cite{Teh01122006}.

Short texts such as tweets pose sparsity challenges for bag-of-words models, which motivates semantic representations based on contextual embeddings.
Embedding-based topic models induce topics by clustering in embedding space, as in Top2Vec \cite{DBLP:journals/corr/abs-2008-09470}, or combine embeddings with sparse lexical representations to obtain interpretable descriptors, as in BERTopic \cite{grootendorst2022bertopicneuraltopicmodeling}, which has also been applied to the analysis of congressional Twitter discourse \cite{Mendonca2024TopicExtraction}.
A related line of work models topics directly in embedding spaces \cite{dieng2020topicmodelinginembeddingspaces} or uses contextualized topic models \cite{bianchi2021pretrainingishottopic}.

\subsection{Dynamic and online topic models}
Dynamic topic models incorporate temporal dependencies, typically encouraging gradual evolution in topic representations. DTM \cite{10.1145/1143844.1143859} uses state-space dynamics over topic parameters; Topics over Time \cite{10.1145/1150402.1150450} models continuous timestamp distributions; and continuous-time variants generalize to irregular time \cite{wang2015continuoustimedynamictopic}. Bayesian nonparametric models such as Timeline target birth and death processes \cite{ahmed2012timelinedynamichierarchicaldirichlet}.

For streaming scenarios, online variants update topics without reprocessing the full history \cite{hoffman2010onlinelearninglda,al-sumait2008onlineLDA}, and surveys summarize topic detection and tracking in social media streams \cite{ibrahim2018surveytoolsapproachestopicdetectiontwitterstreams}. Recent neural dynamic models evolve latent representations (e.g., DETM \cite{dieng2019dynamicembeddedtopicmodel}) and introduce structured dependencies that can support branching/merging behavior \cite{miyamoto-etal-2023-dynamic,cvejoski2023ndftm}. Contrastive and chain-free evolution-tracking formulations further relax one-to-one temporal identities \cite{wu2024chainfree}. Although these approaches capture temporal regularities, discrete split/merge events often remain implicit or require post hoc interpretation.

\subsection{Cluster evolution, split/merge events, and explainability}
Beyond probabilistic modeling, cluster evolution has been studied via similarity graphs and objectives that balance snapshot fit and temporal smoothness, such as evolutionary clustering \cite{teh2006hierarchical}. Visual analytics and alluvial-style representations highlight branching and convergence patterns \cite{rosvall2010mapping}. In practice, split/merge detection is frequently performed by post-processing snapshot clusters with graph-based rules or matching procedures.

BERTilda follows this alignment perspective but adds an explicit, interpretable \emph{flow} constraint: similarity alone can produce ambiguous many-to-many matchings in dense semantic spaces, while document-level coverage helps distinguish substantive transitions from weak topical proximity.

\subsection{Evaluation of topic quality and temporal behavior}
Automated topic evaluation remains challenging. Coherence measures correlate imperfectly with human judgments \cite{chang2009readingtealeavestopicmodels,hoyle2021automatedtopicmodelevaluationbroken}, and different coherence formulations can behave inconsistently across models and domains \cite{roder2015exploringtopiccoherence,ruediger2022topicmodelingrevisited}. Recent work proposes contextualized coherence using masked language models \cite{rahimi2024contextualizedcoherence} and purpose-oriented evaluation enabled by large language models \cite{tan2025purposeoriented}.

For dynamic topic models, evaluation must address temporal behavior. Recent work proposes topic-quality-over-time measures and temporal consistency diagnostics \cite{karakkaparambil-james-etal-2024-evaluating}, and surveys discuss stability and robustness considerations \cite{churchill2022evolutiontopicmodeling,stabilitytopicsurvey2024}.
In this work, we combine within-window topic quality metrics with temporal diagnostics (drift/volatility) and human validation of discrete lifecycle labels.

\section{Method}
\label{sec:method}

\subsection{Overview}
BERTilda has three stages. First, it segments a timestamped corpus into (possibly overlapping) windows and fits a snapshot topic model per window. Second, it aligns topics across adjacent windows by constructing a bipartite graph connecting topics that meet semantic similarity and bidirectional coverage thresholds.
Third, it applies transparent similarity and coverage-based rules to classify lifecycle events (continuation, split, merge, disappearance, or unclear).

\subsection{Windowing and snapshot topics}
Let $\mathcal{D}=\{d_1,\ldots,d_N\}$ be timestamped documents. We define a sequence of windows $t=1,\ldots,T$ with start $s_t$ and end $e_t$:
\begin{align*}
 s_{t+1} &= s_t + \Delta_{\text{step}},\qquad e_t = s_t + \Delta_{\text{window}},\\
 \mathcal{D}_t &= \{d\in\mathcal{D}: s_t\le \text{time}(d) < e_t\}.
\end{align*}
Within each window, a snapshot topic model yields a set of topics $\mathcal{K}_t=\{1,\ldots,K_t\}$ and an assignment $z_t(d)\in\mathcal{K}_t$ for each $d\in\mathcal{D}_t$. Let $\mathcal{D}_{t,k}=\{d\in\mathcal{D}_t:z_t(d)=k\}$ and $n_{t,k}=|\mathcal{D}_{t,k}|$.

\subsection{Topic representations and similarity}
Each document $d$ is embedded as $\phi(d)\in\mathbb{R}^m$ using a fixed encoder shared across windows. Each topic is represented by an embedding $\psi(t,k)\in\mathbb{R}^m$, defined as the centroid of the document embeddings assigned to that topic within the corresponding window. This centroid-based representation ensures topic embeddings remain directly comparable across windows, as they are all derived from a shared document embedding space.

We define topic similarity across adjacent windows via cosine similarity:
\[
\mathrm{sim}((t,k),(t{+}1,j))
= \cos\big(\psi(t,k),\psi(t{+}1,j)\big).
\]

\subsection{Bidirectional coverage by cross-window attribution}
Similarity alone is insufficient to disambiguate many-to-many relations in dense semantic spaces. BERTilda therefore estimates an interpretable \emph{flow} signal by attributing documents across windows.

\paragraph{Forward attribution (outflow).}
For each document $d\in\mathcal{D}_{t,k}$ we select the most similar topic in the next window,
\[
 j^*(d)=\arg\max_{j\in\mathcal{K}_{t+1}} \cos(\phi(d),\psi(t{+}1,j)),
\]
accepting the attribution only if the maximum similarity exceeds $\tau_{\text{doc}}$. We define $F^{\text{fwd}}_t(k\to j)$ as the number of documents in $\mathcal{D}_{t,k}$ attributed to topic $j$ in window $t{+}1$. Outflow coverage is
\[
\mathrm{cov}_{\text{out}}(k\to j)=\frac{F^{\text{fwd}}_t(k\to j)}{n_{t,k}}.
\]

\paragraph{Backward attribution (inflow).}
Symmetrically, for each $d\in\mathcal{D}_{t+1,j}$ we select the most similar topic in the previous window,
\[
 k^*(d)=\arg\max_{k\in\mathcal{K}_t} \cos(\phi(d),\psi(t,k)),
\]
again requiring the maximum similarity to exceed $\tau_{\text{doc}}$.
Let $F^{\text{bwd}}_t(k\to j)$ be the number of documents in $\mathcal{D}_{t+1,j}$ attributed back to topic $k$.
Inflow coverage is
\[
\mathrm{cov}_{\text{in}}(k\to j)=\frac{F^{\text{bwd}}_t(k\to j)}{n_{t+1,j}}.
\]

Intuitively, outflow answers ``where did topic $(t,k)$ go?'', while inflow answers ``where did topic $(t{+}1,j)$ come from?''. This bidirectionality is particularly important for merges: outflow alone can be dominated by large source topics, whereas inflow normalizes by the target topic mass.

\subsection{Event rules and temporal topic graph}

For each adjacent pair of windows $(t,t{+}1)$, we compute cosine similarities between topic centroids and bidirectional coverage by cross-window attribution. We then build a bipartite directed graph from topics in window $t$ to topics in $t{+}1$.
A candidate relation $(k\to j)$ is retained as an edge if it satisfies three minimum requirements,
\begin{equation}
  \mathrm{sim}((t,k),(t{+}1,j)) \ge \tau_{\text{topic}},\qquad
  \mathrm{cov}_{\text{out}}(k\to j) \ge \tau_{\text{cov}}^{\text{out}},\qquad
  \mathrm{cov}_{\text{in}}(k\to j) \ge \tau_{\text{cov}}^{\text{in}}.
  \label{eq:edge_validation}
\end{equation}
Let $E_t$ be the set of retained edges for the pair $(t,t{+}1)$, and define the successor and predecessor sets
\begin{equation}
  S_t(k)=\{j:(k\to j)\in E_t\},\qquad
  P_{t+1}(j)=\{k:(k\to j)\in E_t\}.
  \label{eq:succ_pred_sets}
\end{equation}

We label lifecycle events using simple, auditable rules.
Topic $(t,k)$ is labeled as \emph{continued} if $S_t(k)=\{j\}$ and the unique successor satisfies a stricter similarity-and-coverage condition,
\begin{equation}
  \mathrm{sim}(k,j) \ge \alpha_{\text{cont}},\qquad
  \mathrm{cov}_{\text{out}}(k\to j) \ge \alpha_{\text{cont}}^{\text{out}},\qquad
  \mathrm{cov}_{\text{in}}(k\to j) \ge \alpha_{\text{cont}}^{\text{in}}.
  \label{eq:continued_rule}
\end{equation}

It is labeled as a \emph{split} if it has at least two validated successors and a substantial fraction of its mass flows to them,
\begin{equation}
  |S_t(k)|\ge 2, \quad
  \sum_{j\in S_t(k)} \mathrm{cov}_{\text{out}}(k\to j) \ge \alpha_{\text{split}}^{\text{out}}, \quad
  \mathrm{cov}_{\text{in}}(k\to j) \ge \alpha_{\text{split}}^{\text{in}} \;\; \forall j\in S_t(k).
  \label{eq:split_rule}
\end{equation}

It \emph{disappears} if it has no validated successor, or if the total outflow to all validated successors is negligible,
\begin{equation}
  S_t(k)=\emptyset \quad \text{or} \quad
  \sum_{j\in S_t(k)} \mathrm{cov}_{\text{out}}(k\to j) < \alpha_{\text{disp}}^{\text{out}}
  \quad \text{or} \quad
  \sum_{j\in S_t(k)} \mathrm{cov}_{\text{in}}(k\to j) < \alpha_{\text{disp}}^{\text{in}}.
  \label{eq:disappear_rule}
\end{equation}

Separately, topic $(t{+}1,j)$ is labeled as a \emph{merge target} if it has at least two validated predecessors and receives substantial inflow from them,
\begin{equation}
\begin{aligned}
|P_{t+1}(j)| &\ge 2 \quad, \quad
\sum_{k\in P_{t+1}(j)} \mathrm{cov}_{\text{in}}(k\to j) \ge \alpha_{\text{merge}}^{\text{in}}, \quad \text{and} \\
&\mathrm{cov}_{\text{out}}(k\to j) \ge \alpha_{\text{merge}}^{\text{out}}
\quad \forall k\in P_{t+1}(j).
\end{aligned}
\label{eq:merge_rule}
\end{equation}

Remaining cases are labeled \emph{unclear}. After relations are established and labeled, we construct a directed temporal topic graph whose nodes are time-indexed topics and whose edges correspond to validated cross-window relations.

\subsection{Implementation and complexity notes}
\label{sec:impl}
BERTilda is implemented as a modular pipeline in which the snapshot topic model can be replaced, provided it returns per-window topic assignments and interpretable topic descriptors.
For short texts, we compute document embeddings with a fixed sentence-transformer encoder (all-MiniLM-L6-v2) shared across windows and precompute them once to ensure a common semantic space.
Topic discovery uses BERTopic (transformer embeddings + HDBSCAN clustering + c-TF-IDF descriptors), with a minimum topic size to filter noisy micro-clusters.

For each adjacent window pair, the dominant cost is cross-window attribution, which naively requires comparing each document embedding to all topic embeddings in the neighboring window.
In practice, $K_t$ is typically far smaller than $|\mathcal{D}_t|$, and efficient matrix multiplication makes attribution practical. Candidate edges are additionally pruned by $\tau_{\text{topic}}$, $\tau_{\text{cov}}^{\text{out}}$, and $\tau_{\text{cov}}^{\text{in}}$.

\FloatBarrier
\subsection{Algorithm}
\begin{algorithm}[H]
\caption{BERTilda: temporal topic graph and lifecycle labeling}
\begin{algorithmic}[1]
\REQUIRE Timestamped corpus $\mathcal{D}$; window scheme $(\Delta_{\text{window}},\Delta_{\text{step}})$; thresholds $\tau_{\text{doc}},\tau_{\text{topic}},\tau_{\text{cov}}^{\text{out}},\tau_{\text{cov}}^{\text{in}}$.
\FOR{$t=1$ to $T$}
  \STATE Fit a snapshot topic model on $\mathcal{D}_t$; obtain topics $\mathcal{K}_t$ and representations $\psi(t,\cdot)$.
\ENDFOR
\FOR{$t=1$ to $T-1$}
  \STATE Compute topic similarities $\mathrm{sim}((t,k),(t{+}1,j))$.
  \STATE Compute bidirectional coverage $\mathrm{cov}_{\text{out}}(k\to j)$ and $\mathrm{cov}_{\text{in}}(k\to j)$ by cross-window attribution.
  \STATE Retain edges by Eq.~\ref{eq:edge_validation}; define $S_t(k)$ and $P_{t+1}(j)$ by Eq.~\ref{eq:succ_pred_sets}.
  \STATE Label continuation/split/disappearance/merge/unclear.
\ENDFOR
\end{algorithmic}
\end{algorithm}
\FloatBarrier

\section{Experiments}
\label{sec:exp}

\subsection{Experimental design}
Our experiments are structured around four questions.
First, does the snapshot topic discovery component yield coherent and diverse topics on short political texts, relative to common baselines?
Second, does bidirectional coverage improve the interpretability of split/merge labels relative to minimal alternatives that rely only on topic-to-topic similarity or only on forward flow?
Third, how stable are topic-quality diagnostics over time compared to dynamic topic models with explicit temporal priors?
Fourth, do qualitative case studies illustrate that detected events correspond to meaningful narrative reorganizations?

\subsection{Datasets}
We evaluate on three political corpora.
\emph{Congress tweets}: 357,896 tweets from 544 official accounts of the 119th U.S. Congress (September 26, 2024--September 22, 2025).
\emph{UN General Debates}: annual speeches indexed by year \cite{UNGeneralDebatesKaggle}.
\emph{State of the Union}: presidential addresses indexed by year \cite{StateOfTheUnionCorpusKaggle2025}.

Table~\ref{tab:datasets} summarizes dataset characteristics and the temporal segmentation used for evaluation.

\begin{table}[t]
\centering
\caption{Datasets and temporal segmentation used in experiments.}
\label{tab:datasets}
\begin{tabular}{lllll}
\toprule
Dataset & Domain & Time span & Unit & Windowing \\
\midrule
Congress tweets & social media & 2024--2025 & tweet & 7d window, 3d step \\
UN General Debates & speeches & 1970--2015 & paragraph/speech & 3y window, 2y step \\
State of the Union & speeches & 1790--2018 & paragraph/speech & 7y window, 3y step \\
\bottomrule
\end{tabular}
\end{table}

\subsection{Preprocessing and topic discovery}
We apply standard text normalization (entity decoding, URL/mention cleanup, lemmatization) and retain English-language content.
For tweets, we optionally perform named-entity canonicalization for political actors and organizations.

For snapshot topic discovery, we use BERTopic \cite{grootendorst2022bertopicneuraltopicmodeling} with a transformer encoder (MiniLM) and HDBSCAN clustering. Topics with fewer than 50 documents are discarded (20 for the UN Debates and 1 for the State of the Union). Topic descriptors are defined using the top-$N$ c-TF-IDF words per topic.

\subsection{Baselines and ablations}
\paragraph{Topic discovery baselines.}
We benchmark within-window topic quality against Top2Vec \cite{DBLP:journals/corr/abs-2008-09470} and include bag-of-words baselines (LDA \cite{10.5555/944919.944937} and NMF \cite{lee2001algorithms}) in supplementary material. 

\paragraph{Temporal-context baselines.}
To contextualize stability diagnostics, we compare with DTM \cite{10.1145/1143844.1143859} (Tomotopy implementation) and DETM \cite{dieng2019dynamicembeddedtopicmodel}. Because topic identity differs across modeling families, we interpret drift/volatility as behavioral diagnostics rather than tracking accuracy.

\paragraph{Event-detection baselines (minimal).}
To isolate the contribution of bidirectional coverage and document-level flow, we implement three minimal baselines that operate on the \emph{same} snapshot topics and windows as BERTilda.
(i) \emph{Similarity-only}: build the temporal graph using only $\mathrm{sim}$ and derive pseudo-flow by normalizing similarities per source/target. (ii) \emph{Lexical-only}: replace embedding similarity by c-TF-IDF cosine (or top-word overlap) and normalize analogously. (iii) \emph{Forward-only}: compute outflow by forward attribution only, using forward counts to approximate inflow. Section~\ref{sec:baseline_protocol} details the protocol and how we ensure comparability.

\subsection{Baseline protocol and comparability safeguards}
\label{sec:baseline_protocol}
All event-detection baselines reuse the same preprocessing, windowing, snapshot topics, and (where applicable) topic representations $\psi(t,k)$.
This avoids conflating event detection quality with differences in topic discovery.
For each adjacent window pair, we compute the topic sets $\mathcal{K}_t,\mathcal{K}_{t+1}$ once and store $n_{t,k}$, c-TF-IDF descriptors, and $\psi(t,k)$.

\emph{Similarity-only} computes a similarity matrix $S_{k,j}=\mathrm{sim}((t,k),(t{+}1,j))$ and retains entries above $\tau_{\text{topic}}$.
To make rules comparable to coverage-based thresholds, we derive normalized weights
\[
\widetilde{\mathrm{cov}}_{\text{out}}(k\to j)=\frac{S_{k,j}}{\sum_{j'} S_{k,j'}},\qquad
\widetilde{\mathrm{cov}}_{\text{in}}(k\to j)=\frac{S_{k,j}}{\sum_{k'} S_{k',j}},
\]
and apply the same event rules using $(\widetilde{\mathrm{cov}}_{\text{out}},\widetilde{\mathrm{cov}}_{\text{in}})$. Only the entries of $S_{k,j}$ above the threshold $\tau_{\text{topic}}$ are considered when computing the normalized weights.

\emph{Lexical-only} repeats the procedure using a lexical similarity matrix (c-TF-IDF cosine, or Jaccard overlap of top-$N$ words).

\emph{Forward-only} runs only the forward attribution step, computes $F^{\text{fwd}}_t(k\to j)$, and sets
\[
\mathrm{cov}_{\text{out}}^{\text{fwd}}(k\to j)=\frac{F^{\text{fwd}}_t(k\to j)}{n_{t,k}},\qquad
\mathrm{cov}_{\text{in}}^{\text{fwd}}(k\to j)=\frac{F^{\text{fwd}}_t(k\to j)}{n_{t+1,j}},
\]
thereby approximating inflow without backward attribution.

Finally, all methods share the same default decision thresholds for alignment and event labeling (Table~\ref{tab:hyperparams}) and apply the same post-processing (minimum topic size and candidate-edge pruning).

\subsection{Evaluation metrics}
\label{sec:eval}
We report within-window topic quality using coherence (CV, NPMI, UMass) and diversity; we combine coherence and diversity into an overall topic-quality score following common practice,
\begin{equation}
    \mathrm{TQ}=\mathrm{CV}\times\mathrm{Diversity}.
\end{equation}
Because automated coherence has known failure modes \cite{hoyle2021automatedtopicmodelevaluationbroken}, we treat these metrics as comparative diagnostics rather than definitive quality measures.

To summarize temporal behavior, we compute drift and short-term volatility of topic-quality metrics over time. Drift is estimated by linear regression,
\begin{equation}
  y_t = \beta_0 + \beta_1 t + \varepsilon_t,
\end{equation}
where $\beta_1$ measures drift and significance is tested with a two-sided $t$-test. Volatility is computed as a rolling standard deviation over a window of size $w$.

\subsection{Gold-standard event validation}
\label{sec:gold}
Three independent annotators evaluated a curated subset of detected events from the congressional corpus, balanced across event types. Rather than asking annotators to label events directly (split/merge/continue), we present them with candidate topic correspondences across adjacent windows. For continuation, split, and disappearance items, the focal topic is in window $t$ and annotators see its keywords and representative tweets, together with the three most similar candidate successors in $t{+}1$. For merge items, the focal topic is in $t{+}1$ and annotators see the three most similar candidate predecessors in $t$. Annotators select which candidates are genuinely related, from which we infer event validity. Thirty events per type (excluding unclear) were randomly sampled from BERTilda's output. Validation rates are computed using majority agreement.

\begin{table}[t]
\centering
\small
\caption{Default decision thresholds used in the alignment and event-labeling stages. Dataset-specific window size and step are described in the experimental setup.}
\label{tab:hyperparams}
\begin{tabular}{@{}llc@{}}
\toprule
Symbol & Meaning & Value \\
\midrule
\multicolumn{3}{@{}l}{\textit{Similarity and edge validation}} \\
$\tau_{\text{doc}}$ & min doc-to-topic similarity for attribution & 0.40 \\
$\tau_{\text{topic}}$ & min topic-to-topic similarity for candidate edge & 0.70 \\
$\tau_{\text{cov}}^{\text{out}}=\tau_{\text{cov}}^{\text{in}}$ & min coverage for a validated edge & 0.30 \\
\midrule
\multicolumn{3}{@{}l}{\textit{Event-labeling thresholds}} \\
$\alpha_{\text{cont}}$ & continuation similarity threshold & 0.90 \\
$\alpha_{\text{cont}}^{\text{out}}=\alpha_{\text{cont}}^{\text{in}}$ & min inflow/outflow coverage for continuation & 0.50 \\
$\alpha_{\text{split}}^{\text{out}}$ & min total outflow coverage for split & 0.70 \\
$\alpha_{\text{split}}^{\text{in}}$ & min inflow coverage per successor & 0.50 \\
$\alpha_{\text{disp}}^{\text{out}}=\alpha_{\text{disp}}^{\text{in}}$ & min inflow/outflow coverage for disappearance & 0.40 \\
$\alpha_{\text{merge}}^{\text{out}}$ & min outflow coverage per predecessor & 0.50 \\
$\alpha_{\text{merge}}^{\text{in}}$ & min total inflow coverage for merge target & 0.70 \\
\bottomrule
\end{tabular}
\end{table}

\section{Results}
\label{sec:results}
We first report within-window topic quality and fragmentation on the congressional tweets corpus, then validate event labels on a human-annotated gold set and compare to minimal baselines. Finally, we provide temporal stability diagnostics on historical corpora and illustrate the resulting event narratives via a qualitative case study.

\subsection{Static topic quality and fragmentation (Congress)}
For fair comparison, Top2Vec was configured with the same embedding model (MiniLM), 10 top words per topic, and a minimum of 50 documents per topic.
Across all temporal windows, BERTilda produced 2,569 topics, while Top2Vec produced 5,717 topics, often exceeding 70 topics per week, which is consistent with greater fragmentation.
Table~\ref{tab:congress_metrics} summarizes topic quality metrics.

\begin{table}[t]
\centering
\caption{Static topic quality metrics on the congressional tweets dataset.}
\label{tab:congress_metrics}
\begin{tabular}{lrrrrr}
\toprule
Model & CV & NPMI & UMass & Diversity & TQ \\
\midrule
BERTilda (BERTopic + temporal graph) & 0.7381 & 0.1981 & -2.8489 & 0.9751 & 0.7200 \\
Top2Vec & 0.3565 & $-0.3005$ & $-11.5995$ & 0.8937 & 0.3184 \\
\bottomrule
\end{tabular}
\end{table}

\subsection{Gold-standard event validation (Congress)}
Annotators evaluated a curated subset of 120 detected events from the congressional dataset (30 per event type, sampled from BERTilda outputs). We report annotator-confirmed validation rate, interpreted as precision on detected events, using majority agreement across three annotators.
Table~\ref{tab:event_counts} reports the distribution of event labels predicted by each method on the same 120 annotated items. Table~\ref{tab:event_baselines} reports precision per predicted event type; entries are marked as \emph{n/a} when a method predicts zero events of that type on the annotated set.

\subsection{Baselines and ablations on event detection}
Table~\ref{tab:event_baselines} compares BERTilda with minimal baselines/ablations designed to isolate key design choices. Because all variations operate on the same snapshot topics and windows, performance differences can be attributed to the alignment and labeling mechanism rather than topic discovery.

\begin{table}[t]
\centering
\caption{Distribution of predicted event labels on the 120 annotated gold-set items.}
\label{tab:event_counts}
\begin{tabular}{lcccc}
\toprule
Method & Continue & Disappear & Split & Merge \\
\midrule
BERTilda (similarity + bidirectional coverage) & 30 & 30 & 30 & 30 \\
Similarity-only (normalized topic similarity) & 20 & 48 & 21 & 20 \\
Lexical-only (normalized c-TF-IDF similarity) & 4 & 106 & 0 & 0 \\
Forward-only (no backward attribution) & 20 & 48 & 20 & 21 \\
\bottomrule
\end{tabular}
\end{table}

\begin{table}[t]
\centering
\caption{Annotator-confirmed validation rate (precision) on predicted events, using majority vote across three annotators.}
\label{tab:event_baselines}
\begin{tabular}{lcccc}
\toprule
Method & Continue & Disappear & Split & Merge \\
\midrule
BERTilda (similarity + bidirectional coverage) & 0.867 & 0.800 & 0.767 & 0.667 \\
Similarity-only (normalized topic similarity) & 0.750 & 0.417 & 0.667 & 0.450 \\
Lexical-only (normalized c-TF-IDF similarity) & 0.750 & 0.226 & n/a & n/a \\
Forward-only (no backward attribution) & 0.800 & 0.558 & 0.826 & 0.714 \\
\bottomrule
\end{tabular}
\end{table}

Across all four event types, BERTilda yields the highest macro-average validation rate among the compared methods (0.775), ahead of similarity-only (0.571) and forward-only (0.725). Forward-only remains competitive on split and merge precision, and is slightly higher on those two event types in this subset, but it drops on disappearance, which is consistent with backward attribution helping to normalize inflow and reduce spurious continuations.

\subsection{Robustness to threshold perturbations}
To test whether event validation depends on a narrow hyperparameter choice, we performed a local one-at-a-time sensitivity analysis around the default configuration on the congressional gold set. Preprocessing, windowing, snapshot topics, topic representations, the 120 annotated items, and the majority-vote evaluation protocol were kept fixed, and we varied only one threshold at a time. We focused on the thresholds most directly tied to alignment and labeling, namely $\tau_{\text{topic}}$, $\tau_{\text{doc}}$, $\alpha_{\text{split}}^{\text{out}}$, $\alpha_{\text{merge}}^{\text{in}}$, and $\alpha_{\text{disp}}^{\text{out}}$, and compared BERTilda against forward-only, the strongest ablation.

\begin{figure}[t]
    \centering
    \begin{minipage}{0.49\textwidth}
        \centering
        \includegraphics[width=\linewidth]{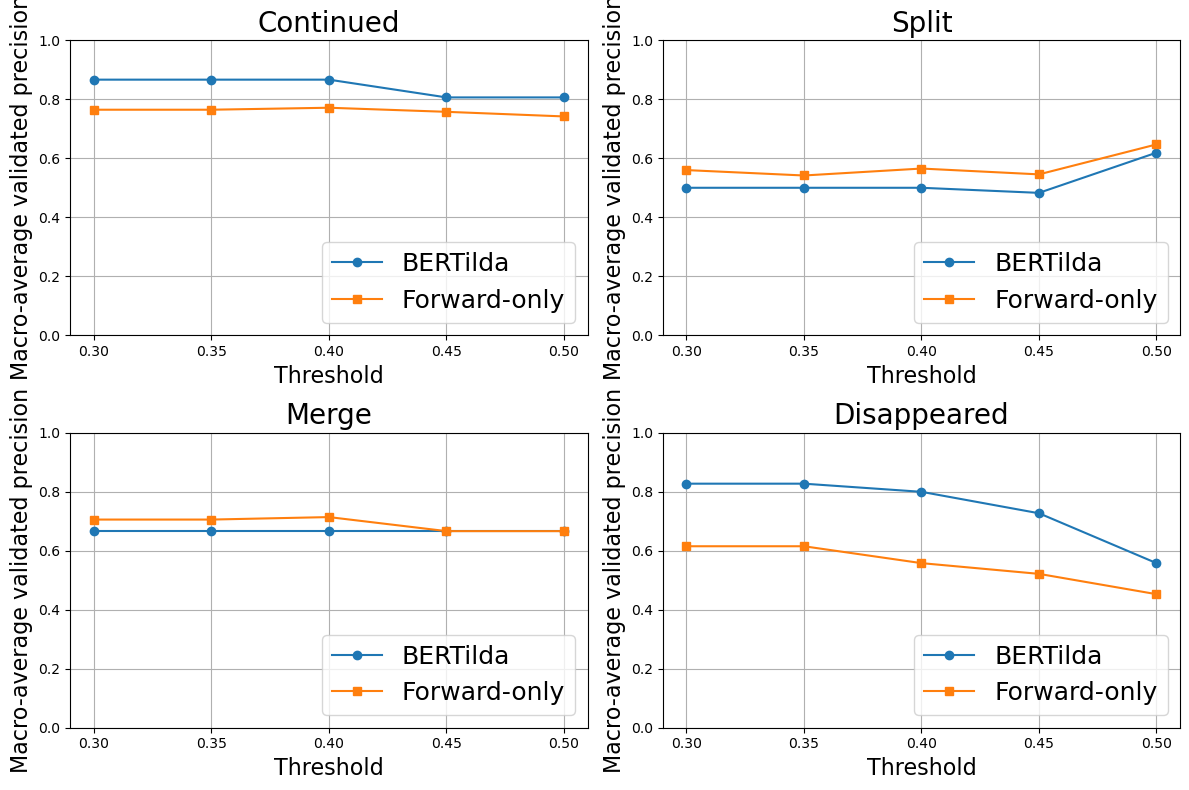}
    \end{minipage}\hfill
    \begin{minipage}{0.49\textwidth}
        \centering
        \includegraphics[width=\linewidth]{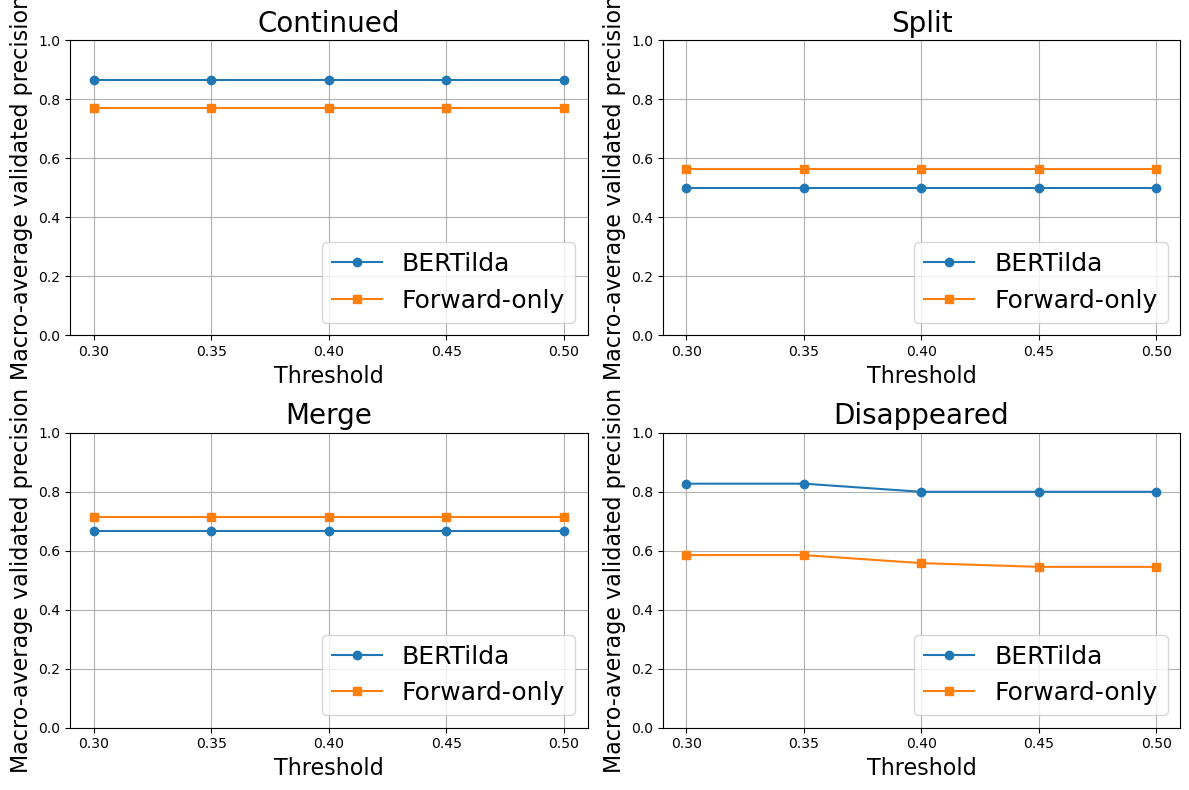}
    \end{minipage}
    \caption{Representative local threshold-sensitivity analyses on the congressional gold set. Left: varying $\tau_{\text{doc}}$. Right: varying $\alpha_{\text{disp}}^{\text{out}}$. Full curves for all five thresholds are reported in the supplementary material.}
    \label{fig:threshold_robustness}
\end{figure}

Across the five perturbation analyses, BERTilda's macro-average validated precision remained between 0.663 and 0.718, whereas forward-only ranged from 0.623 to 0.669. The advantage was most stable for disappearance detection; split and merge were more sensitive, but the method did not collapse into degenerate label distributions under small threshold changes. This supports the claim that BERTilda's gains are not confined to one hand-tuned threshold configuration.

\subsection{Temporal stability diagnostics (historical corpora)}
To contextualize the temporal behavior of within-window topic quality metrics, we report results on the UN General Debates dataset below. Additional State of the Union results are deferred to the supplementary material. Because topic identity differs across model families, these metrics should be read as qualitative points of reference rather than tracking accuracy.

\subsubsection{UN General Debates (1970--2015)}
All texts were preprocessed using a unified pipeline (lowercasing, punctuation/digit removal, lemmatization, stopword removal). Paragraphs were temporally indexed by year. For BERTilda we use 3-year windows with a 2-year step.

\begin{table}[t]
\caption{Overall topic quality metrics on UN General Debates (mean $\pm$ std across windows).}
\label{tab:un_metrics}
\centering
\small
\renewcommand{\arraystretch}{0.95}
\begin{tabular}{@{}l@{\hspace{4pt}}c@{\hspace{4pt}}c@{\hspace{4pt}}c@{\hspace{4pt}}c@{\hspace{4pt}}c@{}}
\toprule
Model & CV & NPMI & UMass & Diversity & TQ \\
\midrule
BERTilda & 0.593 $\pm$ 0.066 & 0.084 $\pm$ 0.044 & -1.105 $\pm$ 0.776 & 0.921 $\pm$ 0.074 & 0.551 $\pm$ 0.095 \\
\shortstack[l]{Tomotopy\\DTM} & 0.361 $\pm$ 0.037 & -0.030 $\pm$ 0.013 & -0.697 $\pm$ 0.614 & 0.589 $\pm$ 0.067 & 0.211 $\pm$ 0.019 \\
DETM & 0.542 $\pm$ 0.012 & -- & -- & 0.998 $\pm$ 0.000 & 0.541 $\pm$ 0.012 \\
\bottomrule
\end{tabular}
\end{table}

\begin{table}[t]
\caption{Temporal drift on UN General Debates (drift = $\beta_1$).}
\label{tab:un_drift}
\centering
\begin{tabular}{lccc}
\toprule
Model & Metric & Drift & p-value \\
\midrule
BERTilda & CV & 0.0007 & 0.5635 \\
 & Diversity & 0.0051 & 0.0062 \\
 & TQ & 0.0031 & 0.1062 \\
Tomotopy DTM & CV & 0.0009 & 0.0204 \\
 & Diversity & 0.0016 & 0.0310 \\
 & TQ & 0.0010 & 2.05e-09 \\
DETM & CV & 2.13e-04 & 0.1215 \\
 & Diversity & -1.53e-07 & 0.9717 \\
 & TQ & 2.13e-04 & 0.1219 \\
\bottomrule
\end{tabular}
\end{table}

\subsection{Qualitative case study (Congress)}
\begin{figure}[t]
    \centering
    \includegraphics[width=0.9\textwidth]{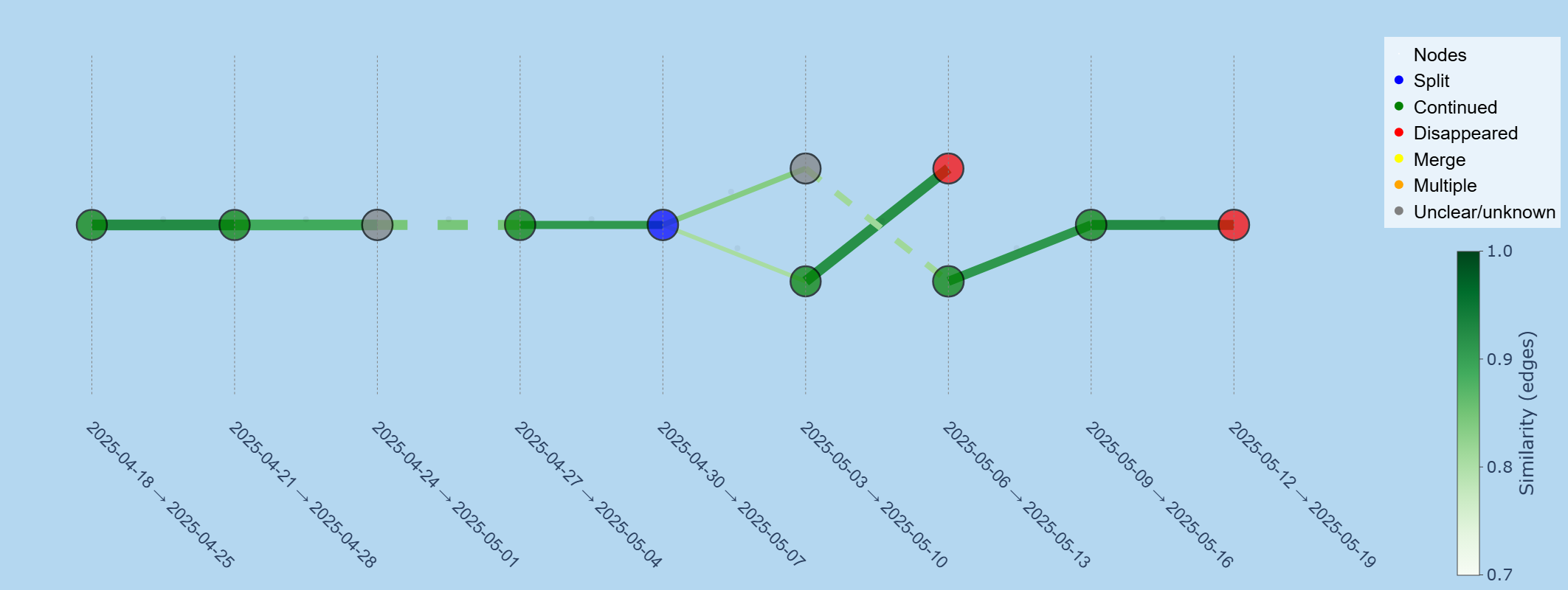}
    \caption{Example temporal evolution of a topic (Holocaust/Israel--Antisemitism), illustrating split, continuations, and disappearance across consecutive windows.}
    \label{fig:HolocaustGraph}
\end{figure}

\emph{Holocaust/Israel--Antisemitism.} From mid-April to mid-May, BERTilda tracks a topic that shifts from Holocaust remembrance to contemporary Israel/Hamas and antisemitism discourse. The system marks this transition as an unclear continuation, followed by a split that separates geopolitical conflict from campus-focused antisemitism debates. The resulting branches then continue independently until disappearance, as illustrated in Figure~\ref{fig:HolocaustGraph}.

\section{Discussion and Limitations}
\label{sec:discussion}

BERTilda provides an explicit representation of topic lifecycles through a temporal topic graph and yields lifecycle labels that are often interpretable under human validation.
In our annotated subset, continuation and disappearance receive the strongest validation rates, while split and merge remain more ambiguous. This is particularly visible for merges, which can reflect partial thematic overlap rather than a single unambiguous fusion.

Several limitations remain.
First, although the local sensitivity analysis suggests that the main conclusions are not tied to a single threshold setting, threshold calibration remains an important open issue, especially for the more ambiguous split and merge cases.
Second, overlapping windows may inflate continuity by construction; ablations with non-overlapping windows or computing flow on window deltas are needed to quantify this effect.
Third, comparisons with DTM/DETM require caution because topic identity is defined differently across modeling families; we treat drift/volatility as behavioral diagnostics rather than tracking correctness.
Finally, our event evaluation focuses on annotator-confirmed precision on detected events; estimating recall would require additional negative sampling beyond the current gold set.

\section{Conclusion}
\label{sec:conclusion}

We presented BERTilda, a graph-based framework for topic lifecycle tracking designed to make split and merge events explicit when topics are discovered independently in each time window.
Our core contribution is an explainable alignment mechanism that combines semantic similarity with bidirectional document flow, supporting explicit successor/predecessor linking and rule-based labeling of continuation, split, merge, disappearance, and unclear transitions.

Empirically, BERTilda achieves competitive topic-quality diagnostics across corpora and yields lifecycle outputs that are often interpretable under human validation.
The ablation-style baselines suggest that document flow, particularly bidirectional coverage, is most helpful for disappearance detection in our setting, while split and merge remain harder and forward-only remains competitive.
Beyond quantitative results, the framework provides an analyst-facing representation (temporal topic graphs) that exposes why an event was labeled as a split or merge, supporting inspection and error analysis.

Future work will focus on three directions.
First, we will conduct systematic robustness analyses over windowing choices and threshold configurations, including calibration against annotated subsets.
Second, we will extend document-to-topic attribution to better handle multi-membership and ambiguity, reducing brittle assignments near cluster boundaries.
Finally, we will broaden the evaluation to additional domains and event types, and explore alternative transport-based or distributional alignment criteria to further improve structural event detection.

\section*{Use of Generative AI}
\label{sec:genai}
Generative AI tools were used to support manuscript preparation in three ways: (i) proofreading and language polishing to improve readability (e.g., grammar correction), (ii) bibliography analysis (e.g., helping identify and organize relevant references), and (iii) formatting and layout assistance (e.g., LaTeX structuring).
All content produced with AI assistance was reviewed and edited by the authors. The authors remain fully responsible for the originality of the manuscript and for the correctness of all technical content, claims, and conclusions.

\begingroup
\small

\section*{Acknowledgments}
This work is funded by national funds through FCT -- Funda\c{c}\~ao para a Ci\^encia e a Tecnologia, I.P., under the support UID/50014/2025 (\url{https://doi.org/10.54499/UID/50014/2025}).

\bibliographystyle{splncs04}
\bibliography{bibliography}

@misc{grootendorst2022bertopicneuraltopicmodeling,
      title={BERTopic: Neural topic modeling with a class-based TF-IDF procedure}, 
      author={Maarten Grootendorst},
      year={2022},
      archivePrefix={arXiv},
      primaryClass={cs.CL}, 
}

@article{10.5555/944919.944937,
author = {Blei, David M. and Ng, Andrew Y. and Jordan, Michael I.},
title = {Latent dirichlet allocation},
year = {2003},
issue_date = {3/1/2003},
publisher = {JMLR.org},
volume = {3},
number = {null},
issn = {1532-4435},
journal = {J. Mach. Learn. Res.},
month = mar,
pages = {993--1022},
numpages = {30}
}

@article{rosvall2010mapping,
  title        = {Mapping Change in Large Networks},
  author       = {Rosvall M, Bergstrom CT},
  year         = {2010},
  journal = {PLoS ONE, 5(1), e8694},
}

@inproceedings{10.1145/1143844.1143859,
author = {Blei, David M. and Lafferty, John D.},
title = {Dynamic topic models},
year = {2006},
isbn = {1595933832},
publisher = {Association for Computing Machinery},
address = {New York, NY, USA},
doi = {10.1145/1143844.1143859},
booktitle = {Proceedings of the 23rd International Conference on Machine Learning},
pages = {113--120},
numpages = {8},
location = {Pittsburgh, Pennsylvania, USA},
series = {ICML '06}
}

@inproceedings{10.1145/1150402.1150450,
author = {Wang, Xuerui and McCallum, Andrew},
title = {Topics over time: a non-Markov continuous-time model of topical trends},
year = {2006},
isbn = {1595933395},
publisher = {Association for Computing Machinery},
address = {New York, NY, USA},
doi = {10.1145/1150402.1150450},
booktitle = {Proceedings of the 12th ACM SIGKDD International Conference on Knowledge Discovery and Data Mining},
pages = {424--433},
numpages = {10},
location = {Philadelphia, PA, USA},
series = {KDD '06}
}

@article{Teh01122006,
author = {Yee Whye Teh and Michael I Jordan and Matthew J Beal and David M Blei},
title = {Hierarchical Dirichlet Processes},
journal = {Journal of the American Statistical Association},
volume = {101},
number = {476},
pages = {1566--1581},
year = {2006},
publisher = {ASA Website},
doi = {10.1198/016214506000000302},
URL = {https://doi.org/10.1198/016214506000000302},
}

@inproceedings{teh2006hierarchical,
author = {Chakrabarti, Deepayan and Kumar, Ravi and Tomkins, Andrew},
title = {Evolutionary clustering},
year = {2006},
isbn = {1595933395},
publisher = {Association for Computing Machinery},
address = {New York, NY, USA},
doi = {10.1145/1150402.1150467},
booktitle = {Proceedings of the 12th ACM SIGKDD International Conference on Knowledge Discovery and Data Mining},
pages = {554--560},
numpages = {7},
location = {Philadelphia, PA, USA},
series = {KDD '06}
}

@inproceedings{karakkaparambil-james-etal-2024-evaluating,
    title = "Evaluating Dynamic Topic Models",
    author = "Karakkaparambil James, Charu  and
      Nagda, Mayank  and
      Haji Ghassemi, Nooshin  and
      Kloft, Marius  and
      Fellenz, Sophie",
    editor = "Ku, Lun-Wei  and
      Martins, Andre  and
      Srikumar, Vivek",
    booktitle = "Proceedings of the 62nd Annual Meeting of the Association for Computational Linguistics (Volume 1: Long Papers)",
    month = aug,
    year = "2024",
    address = "Bangkok, Thailand",
    publisher = "Association for Computational Linguistics",
    url = "https://aclanthology.org/2024.acl-long.11/",
    doi = "10.18653/v1/2024.acl-long.11",
    pages = "160--176"
}

@misc{wang2015continuoustimedynamictopic,
      title={Continuous Time Dynamic Topic Models}, 
      author={Chong Wang and David Blei and David Heckerman},
      year={2015},
      archivePrefix={arXiv},
      primaryClass={cs.IR}, 
}

@misc{ahmed2012timelinedynamichierarchicaldirichlet,
      title={Timeline: A Dynamic Hierarchical Dirichlet Process Model for Recovering Birth/Death and Evolution of Topics in Text Stream}, 
      author={Amr Ahmed and Eric P. Xing},
      year={2012},
      archivePrefix={arXiv},
      primaryClass={cs.IR}, 
}

@inproceedings{miyamoto-etal-2023-dynamic,
    title = "Dynamic Structured Neural Topic Model with Self-Attention Mechanism",
    author = "Miyamoto, Nozomu  and
      Isonuma, Masaru  and
      Takase, Sho  and
      Mori, Junichiro  and
      Sakata, Ichiro",
    editor = "Rogers, Anna  and
      Boyd-Graber, Jordan  and
      Okazaki, Naoaki",
    booktitle = "Findings of the Association for Computational Linguistics: ACL 2023",
    month = jul,
    year = "2023",
    address = "Toronto, Canada",
    publisher = "Association for Computational Linguistics",
    url = "https://aclanthology.org/2023.findings-acl.366/",
    doi = "10.18653/v1/2023.findings-acl.366",
    pages = "5916--5930"
}

@article{DBLP:journals/corr/abs-2008-09470,
  author       = {Dimo Angelov},
  title        = {Top2Vec: Distributed Representations of Topics},
  journal      = {CoRR},
  volume       = {abs/2008.09470},
  year         = {2020},
}

@misc{dieng2019dynamicembeddedtopicmodel,
      title={The Dynamic Embedded Topic Model}, 
      author={Adji B. Dieng and Francisco J. R. Ruiz and David M. Blei},
      year={2019},
      archivePrefix={arXiv},
      primaryClass={cs.CL}, 
}

@misc{UNGeneralDebatesKaggle,
  title        = {UN General Debates Dataset},
  author       = {{United Nations / Kaggle}},
  howpublished = {\url{https://www.kaggle.com/datasets/unitednations/un-general-debates}},
  note         = {Accessed in: 01 Dez 2025}
}

@misc{StateOfTheUnionCorpusKaggle2025,
  title        = {State of the Union Corpus (1790--2018)},
  author       = {{Rachael Tatman / Kaggle}},
  howpublished = {\url{https://www.kaggle.com/datasets/rtatman/state-of-the-union-corpus-1989-2017}},
  note         = {Accessed in: 01 Dez 2025}
}

@inproceedings{hofmann1999plsa,
  author    = {Thomas Hofmann},
  title     = {Probabilistic Latent Semantic Analysis},
  booktitle = {Proceedings of the Fifteenth Conference on Uncertainty in Artificial Intelligence (UAI)},
  year      = {1999},
  pages     = {289--296}
}

@article{lee1999learningparts,
  author  = {Daniel D. Lee and H. Sebastian Seung},
  title   = {Learning the parts of objects by non-negative matrix factorization},
  journal = {Nature},
  year    = {1999},
  volume  = {401},
  number  = {6755},
  pages   = {788--791},
  doi     = {10.1038/44565}
}

@inproceedings{lee2001algorithms,
  author    = {Daniel D. Lee and H. Sebastian Seung},
  title     = {Algorithms for Non-negative Matrix Factorization},
  booktitle = {Advances in Neural Information Processing Systems},
  year      = {2001}
}

@inproceedings{hoffman2010onlinelearninglda,
  author    = {Matthew D. Hoffman and David M. Blei and Francis Bach},
  title     = {Online Learning for Latent Dirichlet Allocation},
  booktitle = {Advances in Neural Information Processing Systems},
  year      = {2010}
}

@inproceedings{al-sumait2008onlineLDA,
  author    = {Loulwah Al{-}Sumait and Daniel Barbar{\'{a}} and Carlotta Domeniconi},
  title     = {On-line {LDA}: Adaptive Topic Models for Mining Text Streams with Applications to Topic Detection and Tracking},
  booktitle = {Proceedings of the 2008 Eighth IEEE International Conference on Data Mining},
  year      = {2008},
  pages     = {3--12},
  doi       = {10.1109/ICDM.2008.140}
}

@article{ibrahim2018surveytoolsapproachestopicdetectiontwitterstreams,
  author  = {Reham Ibrahim and Ahmed Elbagoury and Mohamed Kamel and Farouk Karray},
  title   = {Tools and approaches for topic detection from Twitter streams: survey},
  journal = {Knowledge and Information Systems},
  year    = {2018},
  volume  = {54},
  pages   = {511--539},
  doi     = {10.1007/s10115-017-1081-x}
}

@article{dieng2020topicmodelinginembeddingspaces,
  author  = {Adji B. Dieng and Francisco J. R. Ruiz and David M. Blei},
  title   = {Topic Modeling in Embedding Spaces},
  journal = {Transactions of the Association for Computational Linguistics},
  year    = {2020},
  volume  = {8},
  pages   = {439--453},
  doi     = {10.1162/tacl_a_00325}
}

@inproceedings{bianchi2021pretrainingishottopic,
  title     = {Pre-training is a Hot Topic: Contextualized Document Embeddings Improve Topic Coherence},
  author    = {Bianchi, Federico and Terragni, Silvia and Hovy, Dirk},
  booktitle = {Proceedings of the 59th Annual Meeting of the Association for Computational Linguistics and the 11th International Joint Conference on Natural Language Processing (Volume 2: Short Papers)},
  year      = {2021},
  pages     = {759--766},
  doi       = {10.18653/v1/2021.acl-short.96}
}

@inproceedings{cvejoski2023ndftm,
  title     = {Neural Dynamic Focused Topic Model},
  author    = {Cvejoski, Kostadin and S{\'{a}}nchez, Rams{\'{e}}s J. and Ojeda, C{\'{e}}sar},
  booktitle = {Proceedings of the AAAI Conference on Artificial Intelligence},
  year      = {2023},
}

@inproceedings{wu2024chainfree,
  title     = {Modeling Dynamic Topics in Chain-Free Fashion by Evolution-Tracking Contrastive Learning and Unassociated Word Exclusion},
  author    = {Wu, Xiaobao and Dong, Xinshuai and Pan, Liangming and Nguyen, Thong and Luu, Anh Tuan},
  booktitle = {Findings of the Association for Computational Linguistics: ACL 2024},
  year      = {2024},
  doi       = {10.18653/v1/2024.findings-acl.183}
}

@inproceedings{chang2009readingtealeavestopicmodels,
  author    = {Jonathan Chang and Sean Gerrish and Chong Wang and Jordan L. Boyd{-}Graber and David M. Blei},
  title     = {Reading Tea Leaves: How Humans Interpret Topic Models},
  booktitle = {Advances in Neural Information Processing Systems},
  year      = {2009}
}

@inproceedings{roder2015exploringtopiccoherence,
  author    = {Michael R{"{o}}der and Andreas Both and Alexander Hinneburg},
  title     = {Exploring the Space of Topic Coherence Measures},
  booktitle = {Proceedings of the Eighth ACM International Conference on Web Search and Data Mining},
  year      = {2015},
  pages     = {399--408},
  doi       = {10.1145/2684822.2685324}
}

@inproceedings{hoyle2021automatedtopicmodelevaluationbroken,
  title     = {Is Automated Topic Model Evaluation Broken?: The Incoherence of Coherence},
  author    = {Hoyle, Alexander and Goel, Pranav and Peskov, Denis and Boyd-Graber, Jordan and Resnik, Philip},
  booktitle = {Advances in Neural Information Processing Systems},
  year      = {2021},
}

@article{ruediger2022topicmodelingrevisited,
  author  = {R{"{u}}diger, Moritz and Antons, David and Joshi, Aarti and Salge, Torsten O.},
  title   = {Topic modeling revisited: New evidence on algorithm performance and quality metrics},
  journal = {PLoS ONE},
  year    = {2022},
  volume  = {17},
  number  = {4},
  doi     = {10.1371/journal.pone.0266325}
}

@inproceedings{rahimi2024contextualizedcoherence,
  title     = {Contextualized Topic Coherence Metrics},
  author    = {Rahimi, Hamed and Mimno, David and Hoover, Jacob Louis and Naacke, Hubert and Constantin, Cam{\'{e}}lia and Amann, Bernd},
  booktitle = {Findings of the European Chapter of the Association for Computational Linguistics: EACL 2024},
  year      = {2024},
  pages     = {1760--1773},
  doi       = {10.18653/v1/2024.findings-eacl.123}
}

@article{tan2025purposeoriented,
  author  = {Tan, Zeyi and D'Souza, Jennifer},
  title   = {Toward purpose-oriented topic model evaluation enabled by large language models},
  journal = {International Journal on Digital Libraries},
  year    = {2025},
  volume  = {26},
  doi     = {10.1007/s00799-025-00429-5}
}

@article{churchill2022evolutiontopicmodeling,
  author  = {Churchill, Rob and Singh, Lisa},
  title   = {The Evolution of Topic Modeling},
  journal = {ACM Computing Surveys},
  year    = {2022},
  doi     = {10.1145/3507900}
}

@article{stabilitytopicsurvey2024,
  author  = {Hosseiny Marani, Ali and Baumer, Eric P. S.},
  title   = {A Review of Stability in Topic Modeling: Metrics for Assessing and Techniques for Improving Stability},
  journal = {ACM Computing Surveys},
  year    = {2024},
  doi     = {10.1145/3623269}
}

@article{Mendonca2024TopicExtraction,
  author       = {Mendon{\c c}a, M. and Figueira, {\'A}.},
  title        = {Topic Extraction: BERTopic's Insight into the 117th Congress's Twitterverse},
  journal      = {Informatics},
  year         = {2024},
  volume       = {11},
  number       = {1},
  pages        = {8},
  doi          = {10.3390/informatics11010008},
}
\endgroup

\end{document}

\title{Supplementary Material for \textit{BERTilda}: Explainable Topic Lifecycle Tracking with Split/Merge Detection via Similarity-and-Flow Temporal Graphs}
\titlerunning{BERTilda Supplementary}
\author{Anonymous}
\institute{Institutions}

\maketitle

\section{Additional historical-corpus diagnostics}

This supplement gathers the auxiliary results omitted from the main paper for space reasons. It serves two purposes. First, it reports the historical-corpus diagnostics on the State of the Union (SOTU) collection, complementing the UN General Debates results shown in the main text. Second, it provides additional robustness and validation evidence that supports, but does not replace, the core claims reported in the paper.

Consistent with the UN results reported in the main paper, BERTilda remains competitive on the SOTU corpus and clearly exceeds Tomotopy DTM on coherence-oriented metrics while preserving very high diversity. DETM remains highly diverse, but the overall topic-quality profile is close to BERTilda rather than clearly superior. This matters because the historical corpora provide a markedly different temporal regime from the congressional tweet stream used for the main case study.

The drift values indicate that topic quality on SOTU is not static across time for any model, but the qualitative picture remains aligned with the main paper: BERTilda does not trade off interpretability for temporal tracking, and its behavior remains stable enough to support lifecycle analysis on long historical corpora.

\begin{table}[!b]
\caption{Static topic quality metrics on the congressional tweets dataset.}
\centering
\label{tab:congress_metrics_supp}
\scriptsize
\renewcommand{\arraystretch}{0.9}
\begin{tabular}{@{}l@{\hspace{4pt}}c@{\hspace{4pt}}c@{\hspace{4pt}}c@{\hspace{4pt}}c@{\hspace{4pt}}c@{}}
\toprule
Model & CV & NPMI & UMass & Diversity & TQ \\
\midrule
BERTilda & 0.7381 & 0.1981 & -2.8489 & 0.9751 & 0.7200 \\
NMF & 0.7697 & 0.2155 & -2.1038 & 0.9793 & 0.7539 \\
LDA & 0.5716 & 0.0448 & -4.2256 & 0.9694 & 0.5541 \\
\bottomrule
\end{tabular}
\end{table}

\begin{table}[!b]
\caption{State of the Union: overall topic quality metrics (mean $\pm$ std across windows).}
\label{tab:sou_metrics_supp}
\centering
\scriptsize
\renewcommand{\arraystretch}{0.95}
\begin{tabular}{@{}l@{\hspace{4pt}}c@{\hspace{4pt}}c@{\hspace{4pt}}c@{\hspace{4pt}}c@{\hspace{4pt}}c@{}}
\toprule
Model & CV & NPMI & UMass & Diversity & TQ \\
\midrule
BERTilda & 0.586 $\pm$ 0.081 & 0.006 $\pm$ 0.070 & -5.916 $\pm$ 2.260 & 0.989 $\pm$ 0.007 & 0.580 $\pm$ 0.080 \\
\shortstack[l]{Tomotopy\\DTM} & 0.473 $\pm$ 0.079 & -0.518 $\pm$ 0.086 & -16.344 $\pm$ 3.926 & 0.904 $\pm$ 0.057 & 0.431 $\pm$ 0.094 \\
DETM & 0.550 $\pm$ 0.133 & -- & -- & 0.998 $\pm$ 0.001 & 0.548 $\pm$ 0.132 \\
\bottomrule
\end{tabular}
\end{table}

\begin{table}[!t]
\caption{State of the Union: temporal drift on topic-quality metrics (drift = $\beta_1$).}
\label{tab:sou_drift_supp}
\centering
\scriptsize
\setlength{\tabcolsep}{4pt}
\begin{tabular}{lccc}
\toprule
Model & Metric & Drift & p-value \\
\midrule
BERTilda & CV & -0.001 & 0.000 \\
          & Diversity & 0.000 & 0.100 \\
          & TQ & -0.001 & 0.000 \\
Tomotopy DTM & CV & -0.000 & 0.655 \\
              & Diversity & -0.000 & 0.000 \\
              & TQ & -0.000 & 0.122 \\
DETM & CV & 0.000 & 0.173 \\
     & Diversity & 0.000 & 0.265 \\
     & TQ & 0.000 & 0.169 \\
\bottomrule
\end{tabular}
\end{table}

\FloatBarrier
\section{Additional event-validation results}

The main paper reports gold-set validation using majority agreement. Under the stricter unanimous-vote criterion, inter-annotator agreement remained substantial (Fleiss' $\kappa = 0.61$, Krippendorff's $\alpha = 0.61$), but absolute validation rates decreased, especially for the more ambiguous split and merge cases. This pattern is expected: unanimous vote rewards only the least ambiguous instances and therefore provides a stricter lower bound on event interpretability.

Even under unanimous vote, BERTilda remains strongest on disappearance validation and competitive on the harder split and merge cases. The stricter criterion therefore reinforces the interpretation from the majority-vote analysis in the main paper: the method's main advantage is not that it removes ambiguity altogether, but that it yields a more balanced and interpretable event layer than simpler alignment variants.

\section{Additional robustness results}

We performed a local one-at-a-time sensitivity analysis around the default threshold configuration, keeping preprocessing, windowing, snapshot topics, topic representations, annotated items, and the majority-vote protocol fixed. The main paper shows the two most central perturbations ($\tau_{\text{doc}}$ and $\alpha_{\text{disp}}^{\text{out}}$); here we report the remaining curves together with compact numeric summaries. Across plausible threshold perturbations, BERTilda remained above forward-only in macro-average validated precision, with the most stable relative advantage observed for disappearance detection. Split and merge varied more, which is consistent with their greater ambiguity under human evaluation, but the method did not collapse into a degenerate label distribution under small local changes to the thresholds.

The event-count ranges help explain why macro precision remains informative in this setting. Forward-only tends to produce more disappearances and fewer merges, whereas BERTilda preserves a broader event distribution over the tested thresholds. This is consistent with the main paper's interpretation that bidirectional attribution helps stabilize inflow estimates and reduces collapse into a narrower set of labels.

\begin{table}[!t]
\centering
\caption{Annotator-confirmed validation rate (precision) on predicted events, using unanimous vote across three annotators.}
\label{tab:event_baselines_supp}
\scriptsize
\setlength{\tabcolsep}{4pt}
\begin{tabular}{@{}>{\raggedright\arraybackslash}p{0.53\textwidth}cccc@{}}
\toprule
Method & Continue & Disappear & Split & Merge \\
\midrule
BERTilda (similarity + bidirectional coverage) & 0.533 & 0.567 & 0.367 & 0.400 \\
Similarity-only (normalized topic similarity) & 0.550 & 0.312 & 0.381 & 0.300 \\
Lexical-only (normalized c-TF-IDF similarity) & 0.750 & 0.160 & n/a & n/a \\
Forward-only (no backward attribution) & 0.457 & 0.395 & 0.391 & 0.429 \\
\bottomrule
\end{tabular}
\end{table}

\begin{figure}[!t]
    \centering
    \begin{minipage}{0.485\textwidth}
        \centering
        \includegraphics[width=\linewidth]{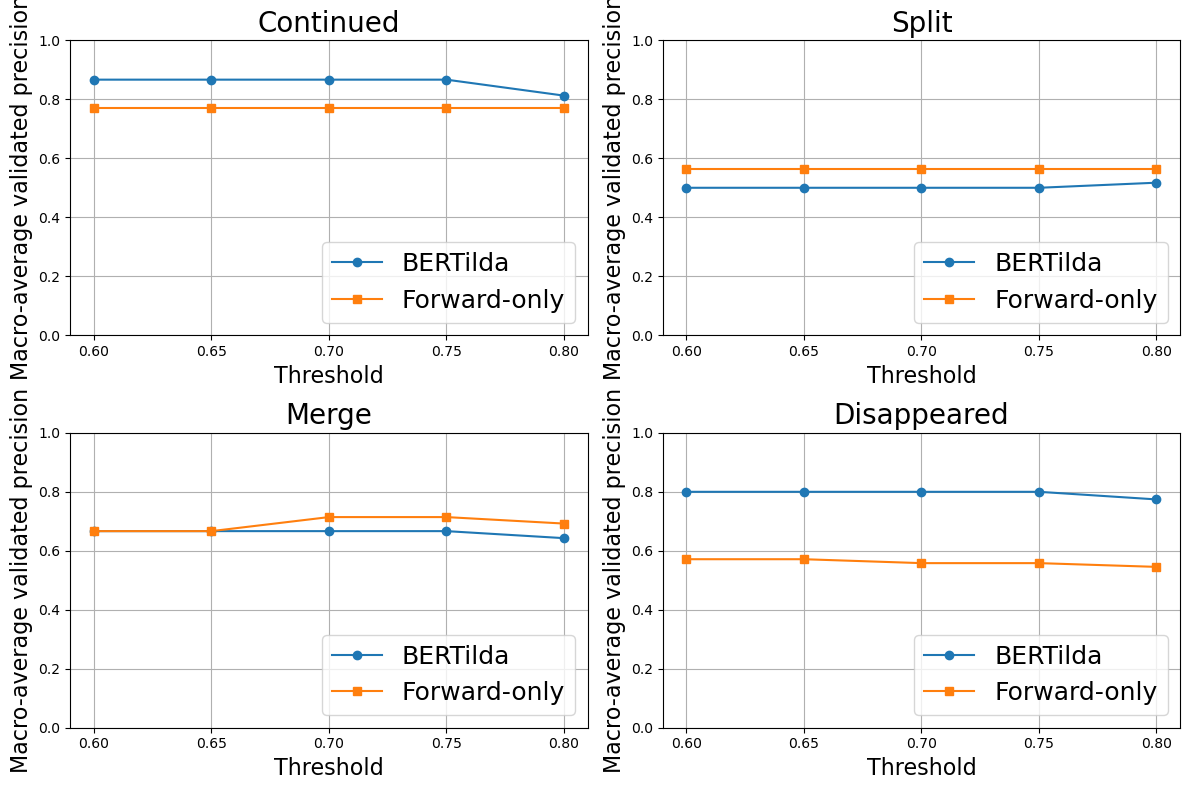}
    \end{minipage}\hfill
    \begin{minipage}{0.485\textwidth}
        \centering
        \includegraphics[width=\linewidth]{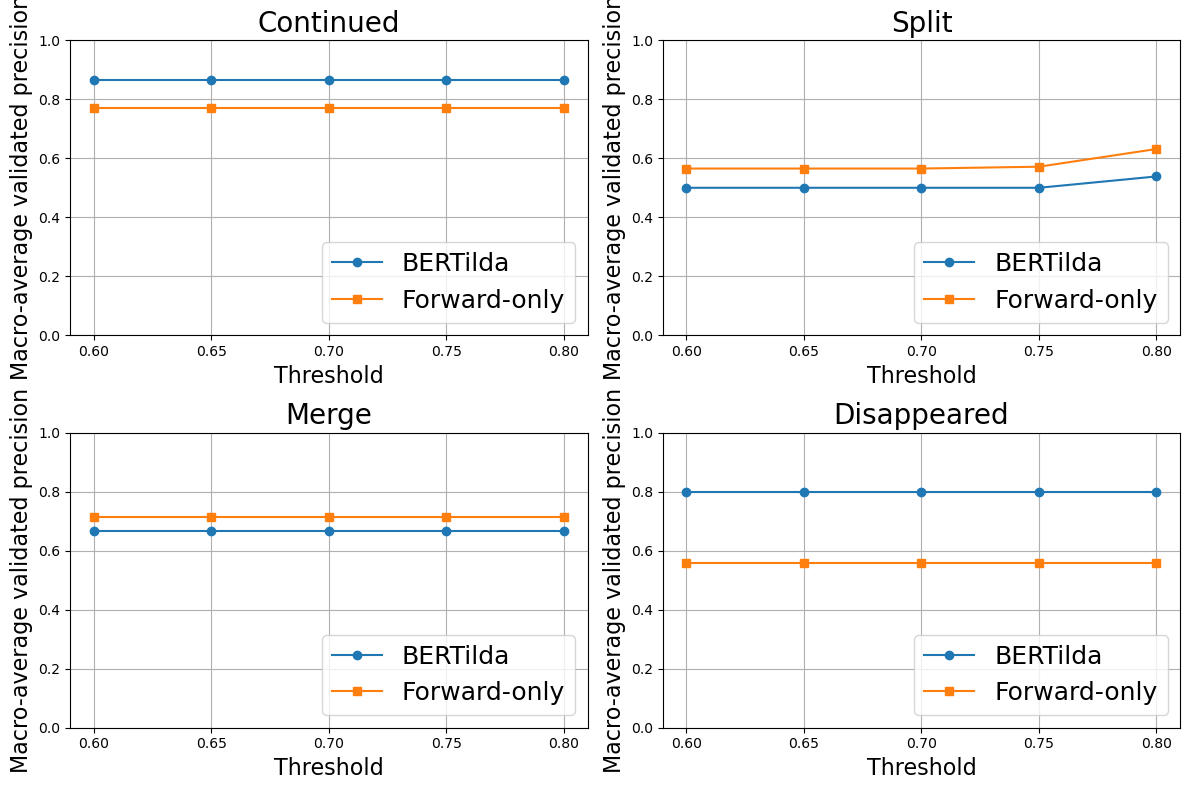}
    \end{minipage}
    \caption{Additional local threshold-robustness curves omitted from the main paper: $\tau_{\text{topic}}$ and $\alpha_{\text{split}}^{\text{out}}$. The remaining perturbation ($\alpha_{\text{merge}}^{\text{in}}$) is summarized numerically in Table~\ref{tab:robustness_summary}.}
    \label{fig:robustness_additional}
\end{figure}

\begin{table}[!t]
\centering
\caption{Macro-average validated precision under local threshold perturbations. Each entry reports min/max/mean/std across the tested values.}
\label{tab:robustness_summary}
\scriptsize
\setlength{\tabcolsep}{4pt}
\begin{tabular}{lcc}
\toprule
Threshold & BERTilda & Forward-only \\
\midrule
$\tau_{\text{topic}}$ & 0.687 / 0.708 / 0.704 / 0.010 & 0.644 / 0.652 / 0.647 / 0.005 \\
$\tau_{\text{doc}}$ & 0.663 / 0.715 / 0.694 / 0.026 & 0.623 / 0.661 / 0.644 / 0.018 \\
$\alpha_{\text{disp}}^{\text{out}}$ & 0.708 / 0.715 / 0.711 / 0.004 & 0.649 / 0.659 / 0.654 / 0.005 \\
$\alpha_{\text{split}}^{\text{out}}$ & 0.708 / 0.718 / 0.710 / 0.004 & 0.652 / 0.669 / 0.656 / 0.007 \\
$\alpha_{\text{merge}}^{\text{in}}$ & 0.683 / 0.708 / 0.701 / 0.011 & 0.624 / 0.652 / 0.643 / 0.013 \\
\bottomrule
\end{tabular}
\end{table}

\begin{table}[!t]
\centering
\caption{Range of predicted event counts across all tested threshold perturbations.}
\label{tab:robustness_event_ranges}
\scriptsize
\setlength{\tabcolsep}{4pt}
\begin{tabular}{lcccc}
\toprule
Method & Continue & Disappear & Split & Merge \\
\midrule
BERTilda & 30--32 & 29--43 & 21--30 & 18--30 \\
Forward-only & 31--35 & 39--53 & 17--25 & 9--17 \\
\bottomrule
\end{tabular}
\end{table}

\clearpage

\section{Additional examples of temporal topic graphs}
To complement the robustness analysis, we include two additional graph examples from the political case studies. These cases illustrate that the graph-based event view remains interpretable across narratives with different temporal granularities and different mixtures of continuation, branching, and disappearance.

\begin{figure}[H]
  \centering
  \begin{minipage}{0.49\textwidth}
    \centering
    \includegraphics[width=\linewidth]{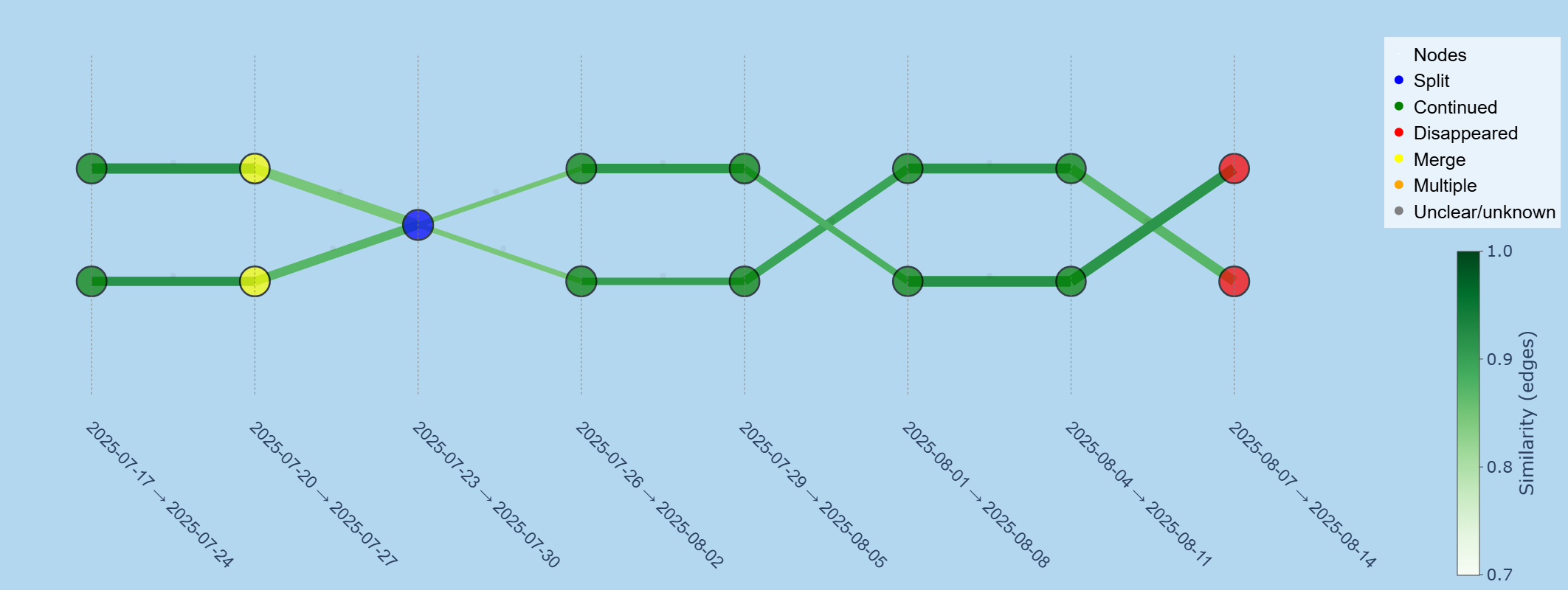}
  \end{minipage}\hfill
  \begin{minipage}{0.49\textwidth}
    \centering
    \includegraphics[width=\linewidth]{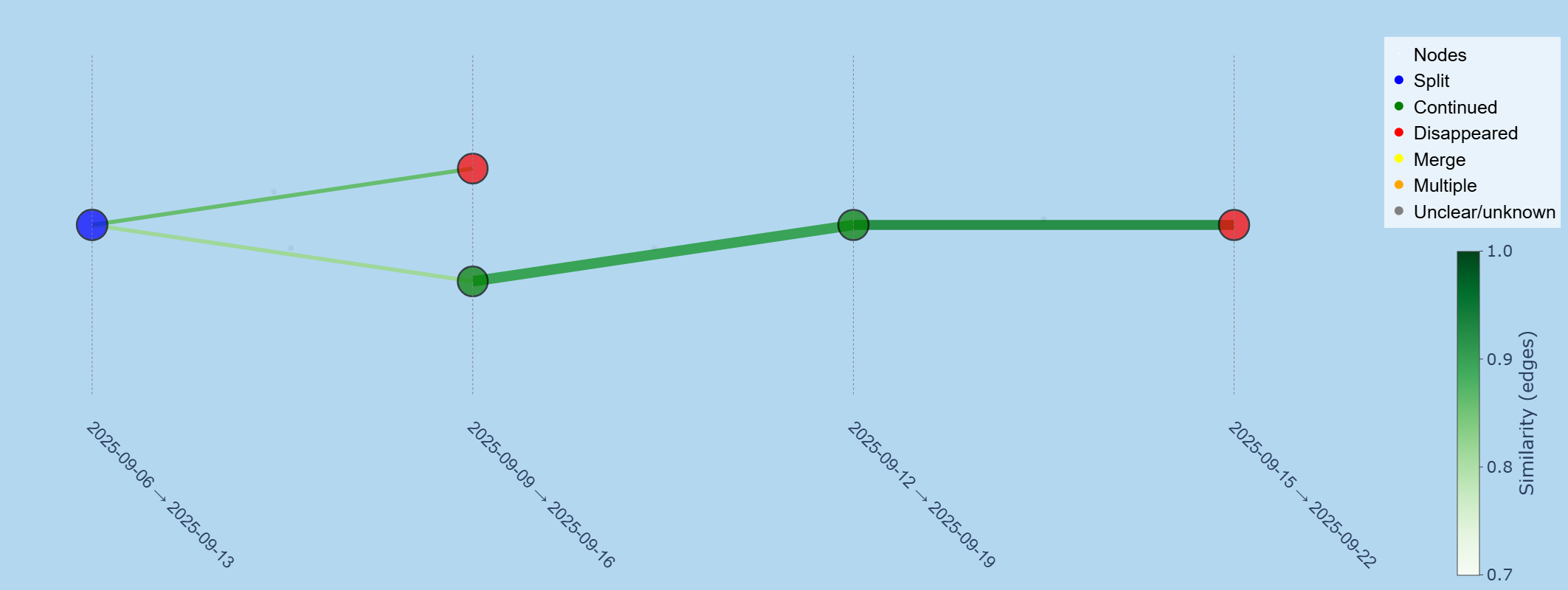}
  \end{minipage}

  \vspace{0.6em}

  \begin{minipage}{0.75\textwidth}
    \centering
    \includegraphics[width=\linewidth]{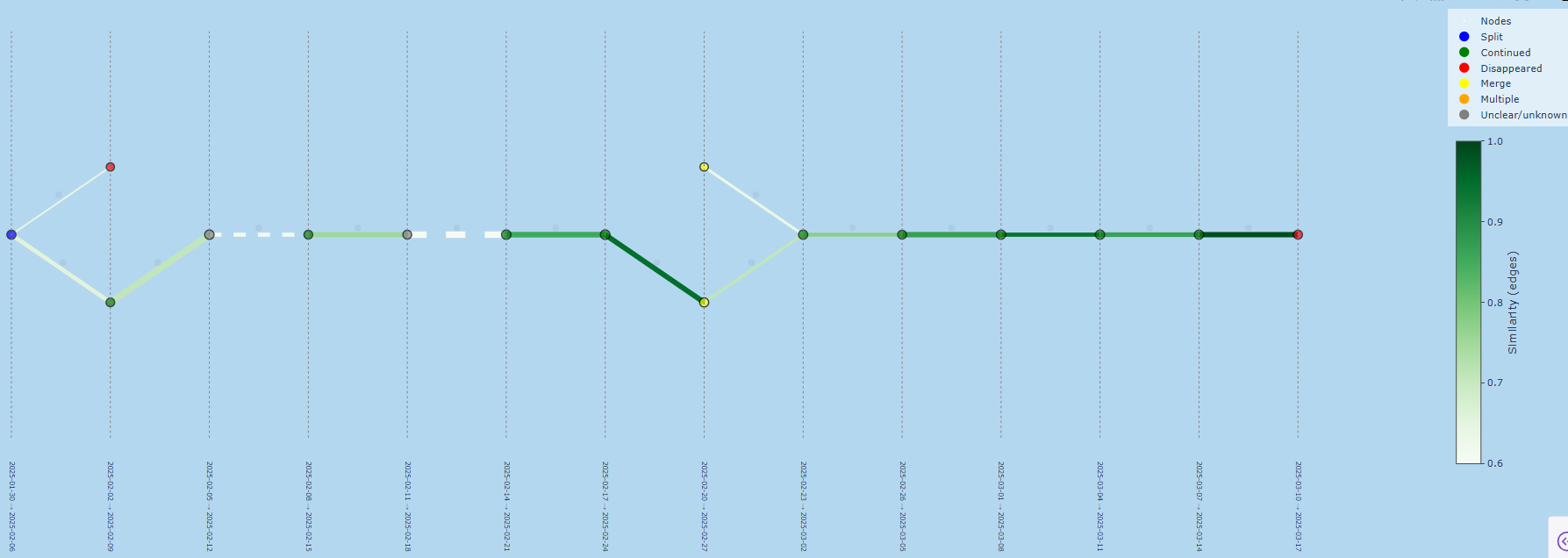}
  \end{minipage}

  \caption{Additional examples of temporal topic graphs from the political case studies: Town Halls and Constituent Services Assistance (top left), Shooting Event to Political Polarization (top right), and Israel-related topic evolution (bottom).}
  \label{fig:extra_topic_graphs}
\end{figure}